# Maximum A Posteriori Inference in Sum-Product Networks

**Jun Mei, Yong Jiang, Kewei Tu**
ShanghaiTech University
{meijun,jiangyong,tukw}@shanghaitech.edu.cn

### Abstract

Sum-product networks (SPNs) are a class of probabilistic graphical models that allow tractable marginal inference. However, the maximum a posteriori (MAP) inference in SPNs is NP-hard. We investigate MAP inference in SPNs from both theoretical and algorithmic perspectives. For the theoretical part, we reduce general MAP inference to its special case without evidence and hidden variables; we also show that it is NP-hard to approximate the MAP problem to $2^{n^\epsilon}$ for fixed $0 \leq \epsilon < 1$, where $n$ is the input size. For the algorithmic part, we first present an exact MAP solver that runs reasonably fast and could handle SPNs with up to 1k variables and 150k arcs in our experiments. We then present a new approximate MAP solver with a good balance between speed and accuracy, and our comprehensive experiments on real-world datasets show that it has better overall performance than existing approximate solvers.

## Introduction

SPNs are a class of probabilistic graphical models known for its tractable marginal inference (Poon and Domingos 2011). In the previous work, SPNs were mainly employed to do marginal inference. On the other hand, although MAP inference is widely used in many applications in natural language processing, computer vision, speech recognition, etc., MAP inference in SPNs has not been widely studied.

Some previous work on MAP inference focuses on selective SPNs (Peharz, Gens, and Domingos 2014), which is also known as determinism in the context of knowledge compilation (Darwiche and Marquis 2002) and arithmetic circuits (Darwiche 2003; Lowd and Domingos 2008; Choi and Darwiche 2017). Huang, Chavira, and Darwiche(2006) presented an exact solver for MAP based on deterministic arithmetic circuits. Peharz et al.(2016) showed that most probable explanation (MPE), a special case of MAP without hidden variables, is tractable on selective SPNs.

Selectivity, however, is not guaranteed in most of the SPN learning algorithms (Gens and Domingos 2012; 2013; Rooshenas and Lowd 2014) and applications (Poon and Domingos 2011; Cheng et al. 2014; Peharz et al. 2014). For SPNs without the selectivity assumption, Peharz(2015) showed that MPE in SPNs is NP-hard by reducing SAT to MPE. Peharz et al.(2016) showed a different proof based on the NP-hardness results from Bayesian networks. Conaty, Mauá, and de Campos(2017) discussed approximation complexity of MAP in SPNs and gave several useful theoretical results.

In this paper, we investigate MAP inference in SPNs from both theoretical and algorithmic perspectives. For the theoretical part, we make the following two contributions. First, we define a special MAP inference problem called MAX that has no evidence and hidden variables, and we show that MAP can be reduced to MAX in linear time. This implies that to study MAP we can instead focus on MAX, which has a much simpler form. Second, we show that it is NP-hard to approximate the MAP problem to $2^{n^\epsilon}$ fox fixed $0 \leq \epsilon < 1$, where $n$ is the input size. This result is similar to a theorem proved by Conaty, Mauá, and de Campos(2017), but we use a proof strategy that is arguably much simpler than theirs. For the algorithmic part, we present an exact MAP solver and an approximate MAP solver. Our comprehensive experiments on real-world datasets show that our exact solver runs reasonably fast and could handle SPNs with up to 1k variables and 150k arcs within ten minutes; our approximate solver provides a good trade-off between speed and accuracy and has better overall performance than previous approximate methods.

## Background

We adapt the notations from Peharz et al.(2015). A random variable is denoted as an upper-case letter, e.g. $X$, $Y$. The corresponding lower-case letter $x$ denotes a value $X$ can assume. The set of all the values $X$ can assume is denoted as $\mathbf{val}(X)$. Thus $x \in \mathbf{val}(X)$.

A set of variables is denoted as a boldface upper-case letter, e.g. $\mathbf{X} = \{X_1, X_2, \ldots, X_N\}$. The corresponding boldface lower-case letter $\mathbf{x}$ denotes a compound value $\mathbf{X}$ can assume. The set of all the compound values $\mathbf{X}$ can assume is denoted as $\mathbf{val}(\mathbf{X})$, i.e. $\mathbf{val}(\mathbf{X}) = \times_{n=1}^{N} \mathbf{val}(X_n)$. Thus $\mathbf{x} \in \mathbf{val}(\mathbf{X})$. For $X \in \mathbf{X}$, $\mathbf{x}[X]$ denotes the projection of $\mathbf{x}$ onto $X$. Thus $\mathbf{x}[X] \in \mathbf{val}(X)$. For $\mathbf{Y} \subseteq \mathbf{X}$, $\mathbf{x}[\mathbf{Y}]$ denotes the projection of $\mathbf{x}$ onto $\mathbf{Y}$. Thus $\mathbf{x}[\mathbf{Y}] \in \mathbf{val}(\mathbf{Y})$.

A compound value $\mathbf{x}$ is also a complete evidence, assigning each variable in $\mathbf{X}$ a value. Partial evidence about $X$ is defined as $\mathcal{X} \subseteq \mathbf{val}(X)$. Partial evidence about $\mathbf{X}$ is defined as $\boldsymbol{\mathcal{X}} := \times_{n=1}^{N} \mathcal{X}_n$. Thus $\boldsymbol{\mathcal{X}} \subseteq \mathbf{val}(\mathbf{X})$. For $X \in \mathbf{X}$, we define $\mathcal{X}[X] := \{\mathbf{x}[X] \mid \mathbf{x} \in \mathcal{X}\}$. Thus $\mathcal{X}[X] \subseteq \mathbf{val}(X)$. For $\mathbf{Y} \subseteq \mathbf{X}$, we define $\mathcal{X}[\mathbf{Y}] := \{\mathbf{x}[\mathbf{Y}] \mid \mathbf{x} \in \mathcal{X}\}$. Thus $\mathcal{X}[\mathbf{Y}] \subseteq \mathbf{val}(\mathbf{Y})$.

## Network polynomials

Darwiche(2003) introduced network polynomials. $\lambda_{X=x} \in \mathbb{R}$ denotes the so-called indicator for $X$ and $x$. $\boldsymbol{\lambda}$ denotes a vector collecting all the indicators of $\mathbf{X}$.

**Definition 1** (Network Polynomial)**.** Let $\Phi$ be an unnormalized distribution over $\mathbf{X}$ with finitely many values. The network polynomial $f_\Phi$ is defined as

$$f_\Phi(\boldsymbol{\lambda}) := \sum_{\mathbf{x} \in \mathbf{val}(\mathbf{X})} \Phi(\mathbf{x}) \prod_{X \in \mathbf{X}} \lambda_{X=\mathbf{x}[X]}. \tag{1}$$

We define $\lambda_{X=x}(\mathbf{x})$ as a function of $\mathbf{x} \in \mathbf{val}(\mathbf{X})$ and $\boldsymbol{\lambda}(\mathbf{x})$ denotes the corresponding vector-valued function, collecting all $\lambda_{X=x}(\mathbf{x})$:

$$\lambda_{X=x}(\mathbf{x}) = \begin{cases} 1 & \text{if } x = \mathbf{x}[X] \\ 0 & \text{otherwise.} \end{cases} \tag{2}$$

It can be easily verified that $f_\Phi(\boldsymbol{\lambda}(\mathbf{x})) = \Phi(\mathbf{x})$ since when we input $\boldsymbol{\lambda}(\mathbf{x})$ to $f_\Phi$, all but one of the terms in the summation evaluate to 0. We extend Eq. 2 to a function of partial evidence $\mathcal{X}$:

$$\lambda_{X=x}(\mathcal{X}) = \begin{cases} 1 & \text{if } x \in \mathcal{X}[X] \\ 0 & \text{otherwise.} \end{cases} \tag{3}$$

Let $\boldsymbol{\lambda}(\mathcal{X})$ be the corresponding vector-valued function. It can also be shown that $f_\Phi(\boldsymbol{\lambda}(\mathcal{X})) = \sum_{\mathbf{x} \in \mathcal{X}} \Phi(\mathbf{x})$, i.e. the network polynomial returns the unnormalized probability measure for partial evidence $\mathcal{X}$. In particular, $f_\Phi(\mathbf{val}(\mathbf{X}))$ returns the normalization constant of $\Phi$.

We should note that, although the indicators are restricted to $\{0, 1\}$ by Eq. 2 and Eq. 3, they are actually real-valued variables. Therefore, taking the first derivative with respect to some $\lambda_{X=x}$ yields

$$\frac{\partial f_\Phi}{\partial \lambda_{X=x}}(\boldsymbol{\lambda}(\mathcal{X})) = \Phi(x, \mathcal{X}[\mathbf{X} \setminus \{X\}]). \tag{4}$$

This means the derivative on the left hand side in Eq. 4 actually evaluates $\Phi$ for modified evidence $\{x\} \times \mathcal{X}[\mathbf{X} \setminus \{X\}]$. This technique will be used in our exact MAP solver.

## Sum-product networks

SPNs over variables with finitely many values are defined as follows:

**Definition 2** (Sum-Product Networks)**.** Let $\mathbf{X}$ be variables with finitely many values and $\boldsymbol{\lambda}$ their indicators. A sum-product network $\mathcal{S} = (\mathcal{G}, \boldsymbol{w})$ over $\mathbf{X}$ is a rooted directed acyclic graph $\mathcal{G} = (V, A)$ with nonnegative parameters $\boldsymbol{w}$. All leaves of $\mathcal{G}$ are indicators and all internal nodes are either sums or products. Denote the set of children of node $\mathsf{N}$ as $\mathbf{ch}(\mathsf{N})$. A sum node $\mathsf{S}$ computes a weighted sum $\mathsf{S}(\boldsymbol{\lambda}) = \sum_{\mathsf{C} \in \mathbf{ch}(\mathsf{S})} w_{\mathsf{S},\mathsf{C}} \mathsf{C}(\boldsymbol{\lambda})$, where the weight $w_{\mathsf{S},\mathsf{C}} \in \boldsymbol{w}$ is associated with the arc $(\mathsf{S}, \mathsf{C}) \in A$. A product node computes $\mathsf{P}(\boldsymbol{\lambda}) = \prod_{\mathsf{C} \in \mathbf{ch}(\mathsf{S})} \mathsf{C}(\boldsymbol{\lambda})$. The output of $\mathcal{S}$ is the function $\mathsf{R}(\boldsymbol{\lambda})$ computed by the root $\mathsf{R}$ and denoted as $\mathcal{S}(\boldsymbol{\lambda})$.

The scope of node $\mathsf{N}$, denoted as $\mathbf{sc}(\mathsf{N})$, is defined as

$$\mathbf{sc}(\mathsf{N}) = \begin{cases} \{X\} & \text{if } \mathsf{N} \text{ is some indicator } \lambda_{X=x} \\ \cup_{\mathsf{C} \in \mathbf{ch}(\mathsf{N})} \mathbf{sc}(\mathsf{C}) & \text{otherwise.} \end{cases} \tag{5}$$

We say an SPN is complete if for each sum $\mathsf{S}$, we have $\mathbf{sc}(\mathsf{C}) = \mathbf{sc}(\mathsf{C}'), \forall \mathsf{C}, \mathsf{C}' \in \mathbf{ch}(\mathsf{S})$. We say an SPN is decomposable if for each product $\mathsf{P}$, we have $\mathbf{sc}(\mathsf{C}) \cap \mathbf{sc}(\mathsf{C}') = \emptyset, \forall \mathsf{C}, \mathsf{C}' \in \mathbf{ch}(\mathsf{P}), \mathsf{C} \neq \mathsf{C}'$. The output function of a complete and decomposable SPN is actually a network polynomial. While there exist SPNs that are not decomposable, in this paper we follow the majority of the previous work and focus on complete and decomposable SPNs.

Now we define MAP inference formally. Using Eq. 2 and Eq. 3, we define $\mathcal{S}(\mathbf{x}) := \mathcal{S}(\boldsymbol{\lambda}(\mathbf{x}))$ and $\mathcal{S}(\mathcal{X}) := \mathcal{S}(\boldsymbol{\lambda}(\mathcal{X}))$. For variables $\mathbf{X}$, we use $\mathbf{Q}, \mathbf{E}, \mathbf{H}$ to denote query, evidence and hidden variables, where $\mathbf{Q} \cup \mathbf{E} \cup \mathbf{H} = \mathbf{X}$, $\mathbf{Q} \neq \emptyset$ and $\mathbf{Q}, \mathbf{E}, \mathbf{H}$ are disjoint. Given $\mathbf{Q}, \mathbf{E}, \mathbf{H}$ and an evidence $\mathbf{e} \in \mathbf{val}(\mathbf{E})$, the MAP inference in the SPN $\mathcal{S}$ over variables $\mathbf{X}$ is defined as

$$MAP_{\mathcal{S}}(\mathbf{Q}, \mathbf{e}, \mathbf{H}) := \arg \max_{\mathbf{q} \in \mathbf{val}(\mathbf{Q})} \mathcal{S}(\{\mathbf{q}\} \times \{\mathbf{e}\} \times \mathbf{val}(\mathbf{H})). \tag{6}$$

Note that MAP inference is typically defined using conditional probabilities, but it is easy to show that our definition is equivalent to the classical definition.

# Theoretical results

## MAX inference

MAP inference splits $\mathbf{X}$ into three parts: query, evidence and hidden variables. We define a special case of MAP inference without evidence and hidden variables, which we call MAX inference:

$$MAX_{\mathcal{S}} := \arg \max_{\mathbf{x} \in \mathbf{val}(\mathbf{X})} \mathcal{S}(\mathbf{x}) \tag{7}$$

We can reduce every MAP problem to a MAX problem in linear time. Without loss of generality, we assume the root of an SPN is a sum (otherwise we can always add a new sum root node linking to the old root with weight 1). Given an SPN $\mathcal{S}$ and a MAP problem with $\mathbf{Q}, \mathbf{e}, \mathbf{H}$, Algorithm 1 modifies $\mathcal{S}$ and returns a new SPN denoted as $\mathcal{S}'$ such that $\forall \mathbf{q} \in \mathbf{val}(\mathbf{Q}), \mathcal{S}'(\mathbf{q}) = \mathcal{S}(\{\mathbf{q}\} \times \{\mathbf{e}\} \times \mathbf{val}(\mathbf{H}))$, which implies $MAX_{\mathcal{S}'} = MAP_{\mathcal{S}}(\mathbf{Q}, \mathbf{e}, \mathbf{H})$. The algorithm runs as follows. We first calculate the value $w_{\mathsf{N}}$ for each node $\mathsf{N}$, which is later multiplied into the arc weights of certain ancestor sum nodes of $\mathsf{N}$. Intuitively, we do bottom-up precomputing of the node values and store the precomputed values in the weights. After that, we remove every node $\mathsf{N}$ and its arcs if $\mathbf{sc}(\mathsf{N}) \subseteq \mathbf{E} \cup \mathbf{H}$ and output the resulting SPN. Using the terminology of knowledge compilation (Darwiche and Marquis 2002) and negation normal forms (Darwiche 2001), the algorithm performs conditioning on the evidence variables, projects the SPN onto the query variables, and then makes simplifications to the SPN structure.

This reduction implies that any efficient algorithm for solving MAX can also be used to efficiently solve MAP. Furthermore, the distribution modeled by $\mathcal{S}'$ is exactly the

**Algorithm 1** Calculate $MAP2MAX_{\mathcal{S}}(\mathbf{Q}, \mathbf{e}, \mathbf{H})$

1: **for all** $\mathsf{N} \in V$ in reverse topological order **do**
2: $\quad w_{\mathsf{N}} \leftarrow 1$
3: $\quad$ **if** $\mathsf{N}$ is a leaf $\lambda_{X=x}$ s.t. $X \in \mathbf{E}$ and $\mathbf{e}[X] \neq x$ **then**
4: $\quad\quad w_{\lambda_{X=x}} \leftarrow 0$
5: $\quad$ **if** $\mathsf{N}$ is a sum $\mathsf{S}$ **then**
6: $\quad\quad$ **for all** $\mathsf{C} \in \mathbf{ch}(\mathsf{S})$ **do**
7: $\quad\quad\quad w_{\mathsf{S,C}} \leftarrow w_{\mathsf{S,C}} w_{\mathsf{C}}$ $\quad \triangleright$ multiply $w_{\mathsf{C}}$ into $w_{\mathsf{S,C}}$
8: $\quad\quad$ **if** $\mathbf{sc}(\mathsf{S}) \subseteq \mathbf{E} \cup \mathbf{H}$ **then**
9: $\quad\quad\quad w_{\mathsf{S}} \leftarrow \sum_{\mathsf{C} \in \mathbf{ch}(\mathsf{S})} w_{\mathsf{S,C}}$ $\quad \triangleright$ otherwise, $w_{\mathsf{S}} = 1$
10: $\quad$ **if** $\mathsf{N}$ is a product $\mathsf{P}$ **then**
11: $\quad\quad w_{\mathsf{P}} \leftarrow \prod_{\mathsf{C} \in \mathsf{P}} w_{\mathsf{C}}$
12: **for all** $\mathsf{N} \in V$ **do**
13: $\quad$ **if** $\mathbf{sc}(\mathsf{N}) \subseteq \mathbf{E} \cup \mathbf{H}$ **then**
14: $\quad\quad$ remove $\mathsf{N}$ and the arcs/weights associated with $\mathsf{N}$

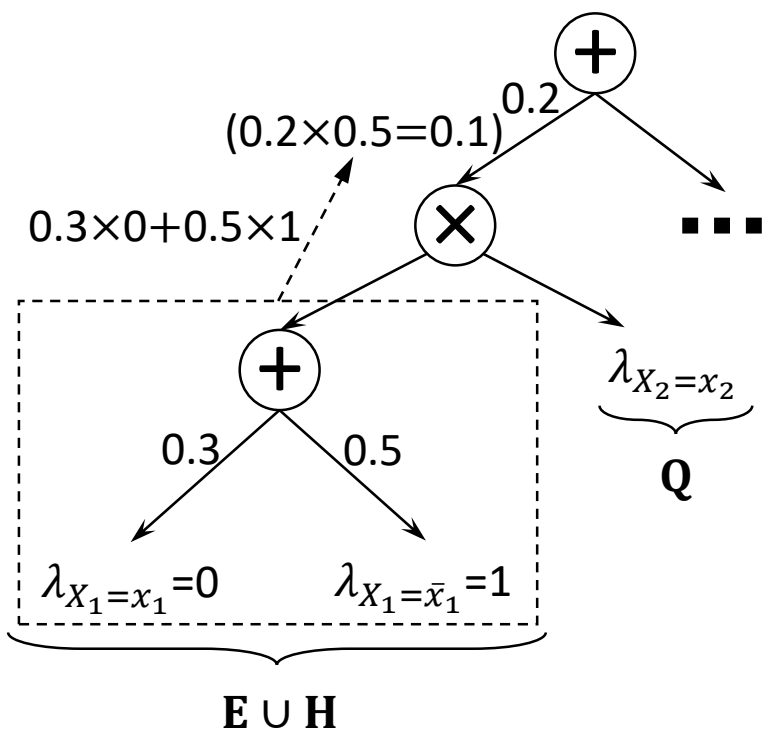

Figure 1: An example of the reduction. $X_1 \in \mathbf{E}$, $\mathbf{e}[X_1] = \bar{x}_1$. The number in the parentheses is the new weight after reduction. Nodes/arcs/weights in the dashed box are removed.

distribution over $\mathbf{Q}$ conditioned on $\mathbf{e}$ modeled by $\mathcal{S}$. Thus, MAP2MAX is an S-reduction (Crescenzi 1997), which implies that any approximation algorithm for MAX can be used to approximate MAP to the same factor. Therefore, in the next two sections we will focus on algorithms solving MAX.

## Approximation complexity

It has been shown in the literature that MAP inference in Bayesian networks (BNs) is hard. Denote the size of an SPN $\mathcal{S}$ and a BN $\mathcal{B}$ as $|\mathcal{S}|$ and $|\mathcal{B}|$ respectively. Theorem 6 in (De Campos 2011) indicates that for any fixed $0 \leq \epsilon < 1$ it is NP-hard to approximate MAP in tree-structured BNs to $2^{|\mathcal{B}|^\epsilon}$. We can transfer this result to SPNs.

**Lemma 1.** *Given a tree-structured BN $\mathcal{B}$, we can construct an SPN $\mathcal{S}$ representing the same distribution with size $|\mathcal{S}| \in \mathcal{O}(|\mathcal{B}|)$ in linear time.*

See the proof of Lemma 1 in the supplementary material.

**Theorem 1.** *For any fixed $0 \leq \delta < 1$, it is NP-hard to approximate MAP in SPNs to $2^{|\mathcal{S}|^\delta}$.*

*Proof.* Suppose there exists fixed $0 \leq \delta < 1$ s.t. it is not NP-hard to approximate MAP in SPNs to $2^{|\mathcal{S}|^\delta}$. Given a tree-structured BN $\mathcal{B}$, we can construct an SPN $\mathcal{S}$ in linear time that represents the same distribution as $\mathcal{B}$. Since $|\mathcal{S}| \in \mathcal{O}(|\mathcal{B}|)$, there exist constants $b, c > 0$ s.t. $|\mathcal{S}| \leq c|\mathcal{B}|$ if $|\mathcal{B}| \geq b$. Define two constants $\tau = (1-\delta)/2$ and $b' = \max\{b, c^{\delta/\tau}\}$. Given a MAP problem in $\mathcal{B}$, we can solve it exactly in constant time if $|\mathcal{B}| < b'$. In the following we consider the case of $|\mathcal{B}| \geq b'$. According to our assumption, we can approximate MAP in the constructed SPN $\mathcal{S}$ to $2^{|\mathcal{S}|^\delta}$ in polynomial time. Since $\mathcal{S}$ and $\mathcal{B}$ represent the same distribution, we have also approximated MAP in $\mathcal{B}$ to $2^{|\mathcal{S}|^\delta}$. Since $|\mathcal{B}| \geq b'$, we have $2^{|\mathcal{S}|^\delta} \leq 2^{(c|\mathcal{B}|)^\delta} \leq 2^{|\mathcal{B}|^\tau |\mathcal{B}|^\delta} = 2^{|\mathcal{B}|^\epsilon}$ where $\epsilon = \delta + \tau < 1$. Therefore, there exists a constant $\epsilon$ s.t. it takes polynomial time to approximate MAP in $\mathcal{B}$ to $2^{|\mathcal{B}|^\epsilon}$, contradicting De Campos's theorem. $\square$

Thm. 1 suggests that it is almost impossible to find a practical and useful approximation bound for MAP inference in SPNs. Note that in parallel to our work, Conaty, Mauá, and de Campos(2017) gave a similar result to Thm. 1 through a reduction from 3-SAT. Both their and our theorems aim at the inapproximability of MAP in SPNs. They suppose the SPNs are trees of low height, which leads to a stronger result than ours. On the other hand, we employ a different proof strategy which is arguably much simpler than theirs.

## Exact solver

Since MAP inference is NP-hard, no efficient exact solver exists (supposing P $\neq$ NP). However, with a combination of pruning, heuristic, and optimization techniques, we can make exact inference reasonably fast in practice. In this section, we introduce two pruning techniques, a heuristic, and an optimization technique in order to build a practical exact solver. We will focus on solving MAX since any MAP problem can be efficiently converted to a MAX problem. Algorithm 2 shows our algorithm framework. The function SEARCH has two augments: $\boldsymbol{\mathcal{X}}$ is the remaining space to be explored and $\mathbf{x}$ is the current best sample.

We first introduce a pruning technique called Marginal Checking (MC). MC computes $\mathcal{S}(\boldsymbol{\mathcal{X}})$ which is the summation of scores of all the samples in $\boldsymbol{\mathcal{X}}$. If it is less than or equal to the score of the current best sample $\mathbf{x}$, then there cannot be any sample in $\boldsymbol{\mathcal{X}}$ with a higher score than $\mathbf{x}$ and therefore we can safely prune space $\boldsymbol{\mathcal{X}}$.

We can go one step further and check and prune the subspaces of $\boldsymbol{\mathcal{X}}$. This leads to a new pruning technique which we call Forward Checking (FC). For each $X \in \mathbf{X}$ and $x \in \boldsymbol{\mathcal{X}}[X]$, we consider the subspace $\{x\} \times \boldsymbol{\mathcal{X}}[\mathbf{X} \setminus \{X\}]$. If the subspace does not have a higher score than $\mathbf{x}$, then we prune the subspace by removing value $x$ from $\boldsymbol{\mathcal{X}}$ (Line 23). The scores of all the subspaces can be computed simultaneously in linear time by taking partial derivatives (Eq. 4). Note that once we prune a subspace by removing a value from $\boldsymbol{\mathcal{X}}$, other subspaces are shrunk and their scores have to be rechecked. For example, the subspace $\{x_1, x_2\} \times \{y\}$ is shrunk to $\{x_2\} \times \{y\}$ if we remove $x_1$. Therefore, we repeat Line 19-23 until $\boldsymbol{\mathcal{X}}$ is no longer changed.

**Algorithm 2** Calculate $\mathbf{x} = MAX_{\mathcal{S}}$

1: $\mathbf{x} \leftarrow$ a initial sample ▷ using any initialization method, for example, random initialization
2: $\mathbf{x} \leftarrow \text{SEARCH}(\mathbf{val}(\mathbf{X}), \mathbf{x})$

3: **function** SEARCH($\boldsymbol{\mathcal{X}}, \mathbf{x}$)
4:     $X \leftarrow$ a variable with $|\boldsymbol{\mathcal{X}}[X]| > 1$ ▷ $|\boldsymbol{\mathcal{X}}[X]| = 1$ means the value of $X$ is determined
5:     **if** no such $X$ exists **then** ▷ all variables are determined
6:         **return** $\mathbf{x}'$ where $\mathbf{x}'$ is the only element in $\boldsymbol{\mathcal{X}}$ ▷ because now $|\boldsymbol{\mathcal{X}}| = 1$ is guaranteed
7:     **for all** $x \in \boldsymbol{\mathcal{X}}[X]$ **do** ▷ consider all possible values of variable $X$
8:         $\boldsymbol{\mathcal{X}}' \leftarrow \{x\} \times \boldsymbol{\mathcal{X}}[\mathbf{X} \setminus \{X\}]$ ▷ new smaller space
9:         $\boldsymbol{\mathcal{X}}' \leftarrow$ MARGINALCHECKING($\boldsymbol{\mathcal{X}}', \mathbf{x}$) or FORWARDCHECKING($\boldsymbol{\mathcal{X}}', \mathbf{x}$)
10:         **if** $\boldsymbol{\mathcal{X}}' \neq \emptyset$ **then**
11:             $\mathbf{x} \leftarrow$ SEARCH($\boldsymbol{\mathcal{X}}', \mathbf{x}$)
12:     **return** $\mathbf{x}$

13: **function** MARGINALCHECKING($\boldsymbol{\mathcal{X}}, \mathbf{x}$)
14:     **if** $\mathcal{S}(\boldsymbol{\mathcal{X}}) > \mathcal{S}(\mathbf{x})$ **then** ▷ check in linear time if there can be better samples than $\mathbf{x}$ in space $\boldsymbol{\mathcal{X}}$
15:         **return** $\boldsymbol{\mathcal{X}}$
16:     **return** $\emptyset$

17: **function** FORWARDCHECKING($\boldsymbol{\mathcal{X}}, \mathbf{x}$)
18:     **repeat**
19:         calculate $\mathbf{D}_x \leftarrow \frac{\partial \mathcal{S}}{\partial x}(\boldsymbol{\mathcal{X}})$ for every $x$ simultaneously ▷ can be done in linear time
20:         **for all** $X \in \mathbf{X}$ **do**
21:             **for all** $x \in \boldsymbol{\mathcal{X}}[X]$ **do**
22:                 **if** $\mathcal{S}(\mathbf{x}) \geq \mathbf{D}_x$ **then** ▷ $\mathcal{S}(\mathbf{x})$ can be cached
23:                     $\boldsymbol{\mathcal{X}} \leftarrow (\boldsymbol{\mathcal{X}}[X] \setminus \{x\}) \times \boldsymbol{\mathcal{X}}[\mathbf{X} \setminus \{X\}]$ ▷ remove $x$ from $\boldsymbol{\mathcal{X}}$
24:     **until** $\boldsymbol{\mathcal{X}}$ is no longer changed
25:     **return** $\boldsymbol{\mathcal{X}}$ ▷ $\boldsymbol{\mathcal{X}}$ is now shrunk and may become $\emptyset$

Now we introduce a heuristic called Ordering, which is inspired by similar techniques for solving constraint satisfaction problems. At Line 4, we need to choose an undetermined variable $X$. Instead of choosing randomly, we choose the variable with the fewest remaining values, i.e., $\arg\min_{X \in \mathbf{X}} |\boldsymbol{\mathcal{X}}[X]|$, which would then lead to fewer search branches. At Line 7, we need to try every value $x \in \boldsymbol{\mathcal{X}}[X]$. We order these values by their corresponding space scores $\mathcal{S}(\{x\} \times \boldsymbol{\mathcal{X}}[\mathbf{X} \setminus \{X\}])$, because we expect a higher score implies that the subspace is more likely to contain a better sample and finding a better sample earlier leads to more effective pruning.

Finally, we introduce an optimization technique called Stage. Once the value of a variable $X$ is determined, it is never changed in the corresponding sub-search-tree. We can treat such determined variables as evidence in MAP inference and reduce the size of the SPN by running Algorithm 1. By doing this, we reduce the amount of computation in the sub-search-tree. Note that, however, the procedure of creating a smaller SPN incurs some overhead. To prevent the overhead from overtaking the benefit, we only do this once every few levels in the search tree.

Since FC is more advanced than MC with similar time complexity, our final exact solver is built by combining FC, Ordering and Stage. Note that our exact solver is actually an anytime algorithm that can terminate at any time and return the current best sample. Thus, our exact solver can also be used as an approximate solver when there is a time budget.

Prior to our work, Huang, Chavira, and Darwiche(2006) also present an exact solver for arithmetic circuits, but their main contribution, a pruning technique (their Algorithm 2), only works on deterministic arithmetic circuits and cannot be easily generalized. In contrast, we focus on more general SPNs without the selectivity (determinism) assumption.

## Approximate solvers

Thm. 1 states that approximating MAP inference in SPNs is very hard. However, in practice it is possible to design approximate solvers with good performance on most data. In this section, we briefly introduce existing approximate methods and then present a new method. Again, when describing the algorithms, we assume the MAP problem has been converted to a MAX problem.

### Existing methods

**Best Tree (BT)** BT, first used by Poon and Domingos(2011), runs in three steps: first, it changes all the sum nodes in the SPN to max nodes; second, it calculates the values of all the nodes from bottom up; third, in a recursive top-down manner starting from the root node, it selects the child of each max node with the largest value. The selected leaf nodes in the third step represent the approximate MAP

**Algorithm 3** Calculate $\hat{\mathbf{x}} = KBT(\mathcal{S})$

1: **for all** $\mathsf{N} \in V$ in reverse topological order **do**
2:     **if** $\mathsf{N}$ is a leaf $\lambda$ **then**
3:         $\mathbf{M}_\lambda \leftarrow \mathbf{best}_K(\{1\})$ ▷ $\mathbf{best}_K(\mathbf{M})$ returns a multiset with at most $K$ best elements in $\mathbf{M}$
4:     **if** $\mathsf{N}$ is a sum $\mathsf{S}$ **then**
5:         $\mathbf{M}_\mathsf{S} \leftarrow \mathbf{best}_K(\uplus_{\mathsf{C} \in \mathbf{ch}(\mathsf{S})}\{w_{\mathsf{S},\mathsf{C}} \times m \mid m \in \mathbf{M}_\mathsf{C}\})$ ▷ in time $\mathcal{O}(|\mathbf{ch}(\mathsf{S})| + K \log |\mathbf{ch}(\mathsf{S})|)$
6:     **if** $\mathsf{N}$ is a product $\mathsf{P}$ **then**
7:         $\mathbf{M}_\mathsf{P} \leftarrow \mathbf{best}_K(\{\prod_{m \in \mathbf{M}'} m \mid \mathbf{M}' \in \times_{\mathsf{C} \in \mathbf{ch}(\mathsf{P})}\mathbf{M}_\mathsf{C}\})$ ▷ in time $\mathcal{O}(K|\mathbf{ch}(\mathsf{P})| \log K)$
8: $\mathbf{S} \leftarrow \{\mathbf{x}$ corresponding to $m \mid m \in \mathbf{M}_\mathsf{R}\}$ ▷ $\mathsf{R}$ is the root; top-down backtracking in time $\mathcal{O}(K|V|)$
9: $\hat{\mathbf{x}} \leftarrow \arg\max_{\mathbf{x} \in \mathbf{S}} \mathcal{S}(\mathbf{x})$ ▷ in time $\mathcal{O}(K|\mathcal{S}|)$

solution of BT. We name this method Best Tree because we can show that it actually finds the *parse tree* of the SPN with the largest value. Tu(2016) showed that any decomposable SPN can be seen as a stochastic context-free And-Or grammar, and following their work we can define a parse tree of an SPN as follows.

**Definition 3** (Parse Tree)**.** Given an SPN $\mathcal{S} = (\mathcal{G}, \boldsymbol{w})$, a parse tree $\mathcal{T} = (\mathcal{G}', \boldsymbol{w}')$ is an SPN where $\mathcal{G}' = (V', A')$ is a subgraph of $\mathcal{G}$ and $\boldsymbol{w}'$ is the subset of $\boldsymbol{w}$ containing weights of the arcs in $A'$. $\mathcal{G}'$ is recursively constructed as follows: 1) we add the root $\mathsf{R}$ of $\mathcal{G}$ into $V'$; 2) when a sum $\mathsf{S}$ is added into $V'$, add exactly one of its children $\mathsf{C}$ into $V'$ and the corresponding arc $(\mathsf{S}, \mathsf{C})$ into $A'$; 3) when a product $\mathsf{P}$ is added into $V'$, add all its children to $V'$ and all the corresponding arcs to $A'$. The value of the parse tree is the product of the weights in $\boldsymbol{w}'$.

The notion of parse trees has been used before in the SPN and arithmetic circuit literature under different terms, e.g., *induced trees* in (Zhao, Poupart, and Gordon 2016). We use the term "parse trees" because our approximate solver is inspired by the formal grammar literature.

**Normalized Greedy Selection (NG)** NG was also used first by Poon and Domingos(2011). It is very similar to BT except that in the first step, NG does not change sum nodes to max nodes. We name this method Normalized Greedy Selection because it can be seen as greedily constructing a parse tree in a recursive top-down manner by selecting for each sum node the child with the largest weight in the locally normalized SPN (Peharz et al. 2015).

**Argmax-Product (AMAP)** AMAP was proposed by Conaty, Mauá, and de Campos(2017). It does $|\mathbf{ch}(\mathsf{S})|$ times bottom-up evaluation on every sum $\mathsf{S}$ in the SPN, so it has quadratic time complexity, while BT and NG both have linear time complexity.

**Beam Search (BS)** Hill climbing has been used in MAP inference of arithmetic circuits (Park 2002; Darwiche 2003), a type of models closely related to SPNs. BS is an extension of hill climbing with $K$ samples. In each round, it evaluates all the samples that result from changing the value of one variable in the existing samples, and then it keeps the top $K$ samples. The evaluation of all such samples in each round can be done in linear time using Eq. 4. The number $K$ is called the beam size.

## K-Best Tree method

It can be shown that the set of leaves of a parse tree $\mathcal{T}$ (Def. 3) corresponds to a single sample $\mathbf{x}$. We denote this relation as $\mathcal{T} \sim \mathbf{x}$. On the other hand, a sample may correspond to more than one parse tree. Formally, $\forall \mathcal{T} \sim \mathbf{x}$, we have $\mathcal{T}(\mathbf{x}) = \mathcal{T}(\mathbf{val}(\mathbf{X}))$ and $\mathcal{T}(\mathbf{x}) \leq \mathcal{S}(\mathbf{x})$. Furthermore, $\mathcal{S}(\mathbf{x}) = \sum_{\mathcal{T} \sim \mathbf{x}} \mathcal{T}(\mathbf{x})$ (Zhao, Poupart, and Gordon 2016). We say a sample is ambiguous with respect to an SPN if it corresponds to more than one parse tree of the SPN. We say an SPN is ambiguous if there exist some ambiguous samples with respect to the SPN. Non-ambiguity is also known as selectivity in (Peharz, Gens, and Domingos 2014). Recall that BT finds the sample with the best parse tree of the SPN. It is easy to show that BT finds the exact solution to the MAX problem if the SPN is unambiguous (Peharz et al. 2016). However, BT cannot find good solutions if the input SPN is very ambiguous, as will be shown in our experiments.

Here we propose an extension of BT called $K$-Best Tree (KBT) that finds the top $K$ parse trees with the largest values (Algorithm 3). KBT is motivated by our empirical finding that even for ambiguous SPNs, in many cases the exact MAX solution corresponds to at least one parse tree with a large (although not necessarily the largest) value. If at least one parse tree of the exact solution is among the top $K$ parse trees, KBT will be able to find the exact solution. Note that the $K$ best trees that KBT finds may not correspond to $K$ unique samples, since there may exist different parse trees corresponding to the same sample.

Similar to BT, KBT runs in two steps. In the bottom-up step, we calculate $K$ best subtrees rooted at each node. In the top-down step, we backtrack to find the $K$ samples corresponding to the $K$ best trees. After that, we evaluate the $K$ samples on the SPN and return the best one. When we set $K = 1$, KBT reduces to BT. Notice that the set notation in Algorithm 3 denotes multisets.

Now we analyze the time complexity of KBT. To execute Line 5, we first push the best value in the multiset of every child into a priority queue and then pop $K$ times. Whenever we pop a value $m$, we push into the queue the next best value (if one exists) in the multiset of the child that $m$ belongs to. The size of the queue is $|\mathbf{ch}(\mathsf{S})|$. The number of pushing is $|\mathbf{ch}(\mathsf{S})| + K$ and the number of popping is $K$. The time complexity is therefore $\mathcal{O}(|\mathbf{ch}(\mathsf{S})| + K \log |\mathbf{ch}(\mathsf{S})|)$ if we use Fibonacci heap as the priority queue.

To execute Line 7, we keep performing pairwise merging of the multisets of the children until we get a single multiset left. When merging two multisets, we first push into a priority queue the product of the best values from the two multisets and then pop $K$ times. Whenever we pop a product $m_1 \times m_2$, we push into the queue two new products $m_1' \times m_2$ and $m_1 \times m_2'$ if we have not pushed them, where $m_1'$ and $m_2'$ are the next best values after $m_1$ and $m_2$ in the two multisets respectively. Thus when merging two multisets, we pop for at most $K$ times and push for $2K+1$ times. We merge $|\mathbf{ch}(\mathsf{P})|-1$ times. Therefore, the time complexity is $\mathcal{O}(K|\mathbf{ch}(\mathsf{P})| \log K)$ if using Fibonacci heap.

Overall, the time complexity of KBT is $\mathcal{O}(|\mathcal{S}|K \log K)$. When $K$ is a constant, the time complexity is linear in the SPN size. There is a trade-off between the running time and accuracy of the algorithm. A large $K$ would likely improve the quality of the result but would lead to more running time.

## Experiments

We evaluated the MAP solvers on twenty widely-used real-world datasets (collected from applications and data sources such as click-through logs, plant habitats, collaborative filtering, etc.) from (Gens and Domingos 2013), with variable numbers ranging from 16 to 1556. We used the LearnSPN method (Gens and Domingos 2013) to obtain an SPN for each dataset. The numbers of arcs of the learned SPNs range from 6471 to 2,598,116. The detailed statistics of the learned SPNs are shown in the supplementary material. We generated MAP problems with different proportions of query (Q), evidence (E) and hidden (H) variables. For each dataset and each proportion, we generated 1000 different MAP problems by randomly dividing the variables into Q/E/H variables. When running the solvers, we bounded the running time for one MAP problem by 10 minutes. We ran our experiments on Intel(R) Xeon(R) CPU E5-2697 v4 @ 2.30GHz. Our code is available at `https://github.com/shtechair/maxspn`.

### Exact solver

We evaluated four combinations of the techniques that we introduced earlier: Marginal Checking (MC), Forward Checking (FC), FC with Ordering (FC+O), and FC with both Ordering and Stage (FC+O+S).

Figure 2 shows, for each dataset and with the Q/E/H proportion being 3/3/4, the number of times each method finished running within 10 minutes on the 1000 problems. Results for additional Q/E/H proportions can be found in the supplementary material. It can be seen that for the datasets with the smallest variable numbers and SPN sizes, all four methods finished running within ten minutes. On the other datasets, FC clearly beats MC and adding Ordering and Staging brings further improvement. Our best method, FC+O+S, can be seen to handle SPNs with up to 1556 variables ("Ad") and 147,599 arcs ("Accidents").

The last four columns of Figure 4 show the average running time of the four methods (with a 10-minute time limit for each problem). On the first three datasets, which have very small variable numbers and SPN sizes, MC is actually faster than the other three methods. This is most likely because on these datasets the overhead of FC and the two additional techniques dominates the running time. On the other datasets, the benefit of FC and the two techniques can be clearly observed.

### Approximate solver

We evaluated all the approximate solvers that we have discussed, as well as the approximate versions of our exact solvers. For BS, we tested beam sizes of 1, 10 and 100. For KBT, we tested $K$ = 10 and 100. We measure the performance of a solver on each dataset with each Q/E/H proportion by its running time and winning count. The winning count is defined as the number of problems on which the solver outputs a solution with the highest score among all the solvers. Since our exact solvers are anytime algorithms, we also evaluated them as approximate solvers with a 10-minute time budget.

Figure 3 shows, for each method and each Q/E/H proportion, the average running time vs. the winning counts averaged over all the datasets. We can see from the figure that the best-tree based methods, BT (=KBT1), KBT10 and KBT100, dominate the other methods with less running time and higher winning counts. Increasing $K$ with KBT improves winning counts but slows down the solver, as one would expect. In terms of running time, BT and NG are much faster than the other methods, while (FC+O+S), the approximate version of the exact solver, is by far the slowest. KBT100 clearly has the highest winning counts, followed by (FC+O+S), KBT10 and AMAP. Furthermore, we see that with the proportion of hidden variables increasing, the winning counts of most methods (except AMAP, KBT100 and KBT10) fall significantly. We believe this is because with more hidden variables, the MAP problem becomes more difficult, as reflected by the fact that the reduced SPN from Algorithm 1 becomes exponentially more ambiguous.

Figure 4 and 5 show the running time and winning counts of all the methods on each dataset under the Q/E/H proportion of 3/3/4. The figures for additional Q/E/H proportions can be found in the supplementary material. We can see that AMAP failed to produce any result within ten minutes on the "20 Newsgroup" dataset, and on the other 19 datasets it actually has higher winning counts but significantly longer running time than KBT100. For the approximate versions of the exact solvers, we can see that even when they were terminated before they could finish, (FC+O) and (FC+O+S) still achieve competitive winning counts, which is in sharp contrast to (FC). This suggests that Ordering is very effective in guiding the search towards good solutions earlier.

While our experimental results are based on a 10-minute time budget, we find that changing the time budget to 2 minutes or 50 minutes leads to no significant change to the results. With a time budget of 2 minutes, the numbers in Figure 4 and 5 will not change if the running time (Figure 4) is well below 120. That means for the eight approximate solvers, only a few numbers of BS100 and AMAP will change (with a new running time of 120 and worse winning counts), and the numbers of the other approximate solvers will not be influenced. On the other hand, the winning counts for the four approximate versions of the exact solvers would likely decrease on many of the datasets. With a time budget of 50 minutes, the numbers in Figure 4 and 5 will not change if the running time (Figure 4) is well below 600. That means for the eight approximate solvers, even fewer numbers of BS100 and AMAP will change (with a new running time of 3000 and potentially better winning counts), and the numbers of the other approximate solvers will not be influenced. We actually find that on the "20 Newsgroup" dataset, AMAP fails to terminate even after 50 minutes, so its winning counts would have no change.

| Dataset | MC | FC | FC+O | FC+O+S |
|---|---|---|---|---|
| NLTCS | 1000 | 1000 | 1000 | 1000 |
| MSNBC | 1000 | 1000 | 1000 | 1000 |
| KDDCup 2k | 1000 | 1000 | 1000 | 1000 |
| Plants | 1000 | 1000 | 1000 | 1000 |
| Audio | 242 | 298 | 354 | 414 |
| Netflix | 39 | 104 | 272 | 340 |
| Jester | 38 | 68 | 178 | 201 |
| Accidents | 1000 | 1000 | 1000 | 1000 |
| Tretail | 3 | 5 | 8 | 8 |
| Pumsb_star | 957 | 1000 | 1000 | 1000 |
| DNA | 24 | 85 | 133 | 160 |
| Ad | 0 | 373 | 970 | 980 |

Figure 2: Finishing counts of exact solvers. We skip the rows of all zeros.

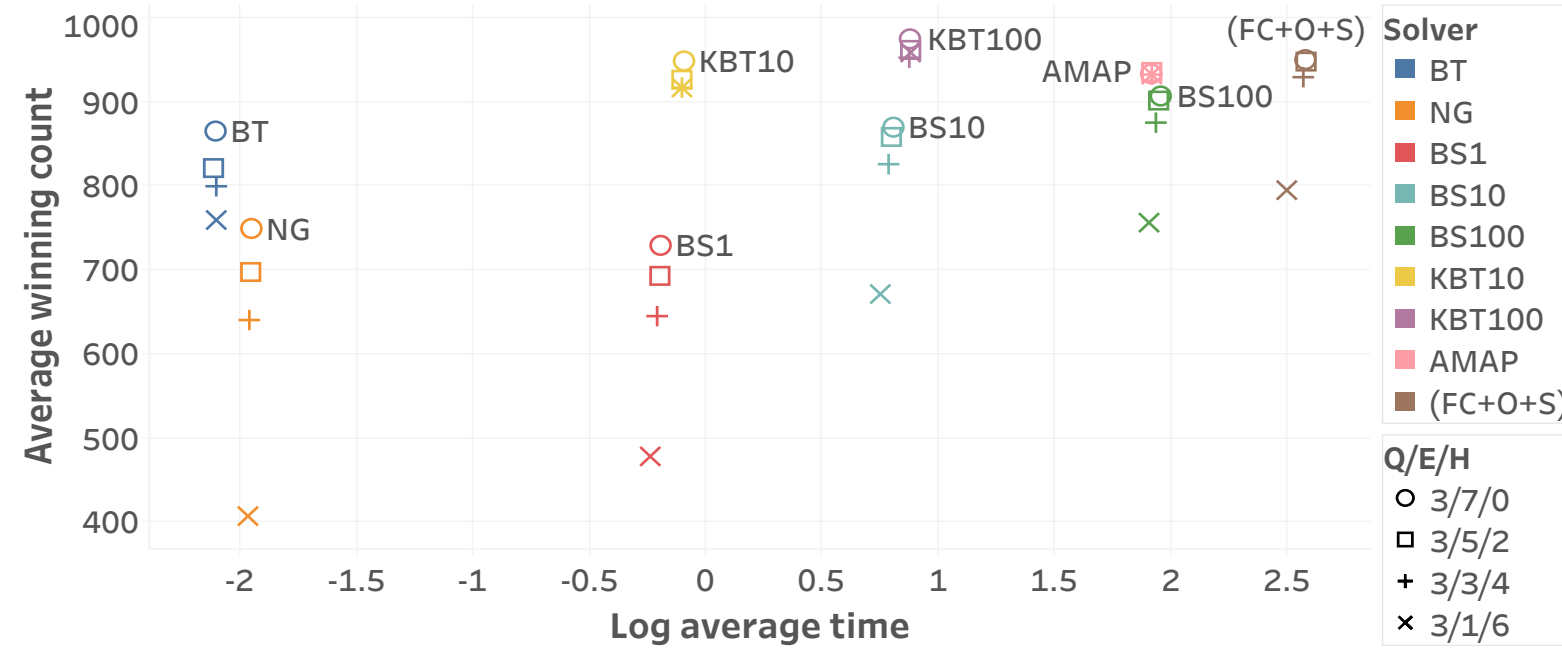

Figure 3: Average running time vs. winning counts averaged over all the datasets. On one of the datasets, AMAP timed out and hence its winning count is set to 0. See discussion in the main text.

| Dataset | BT | NG | BS1 | BS10 | BS100 | KBT10 | KBT100 | AMAP | (MC) | (FC) | (FC+O) | (FC+O+S) |
|---|---|---|---|---|---|---|---|---|---|---|---|---|
| NLTCS | 0.0002 | 0.0004 | 0.0013 | 0.0085 | 0.0302 | 0.0149 | 0.0404 | 0.0587 | 0.0052 | 0.0098 | 0.0085 | 0.0065 |
| MSNBC | 0.0017 | 0.0023 | 0.0091 | 0.0722 | 0.3139 | 0.0945 | 0.3031 | 3.8355 | 0.0175 | 0.0419 | 0.0390 | 0.0316 |
| KDDCup 2k | 0.0030 | 0.0043 | 0.0483 | 0.3963 | 4.0078 | 0.3089 | 2.3278 | 10.368 | 9.6917 | 11.629 | 11.267 | 7.9157 |
| Plants | 0.0026 | 0.0038 | 0.0280 | 0.2482 | 2.6587 | 0.2271 | 1.6251 | 7.9273 | 2.2650 | 1.8247 | 0.7866 | 0.6578 |
| Audio | 0.0025 | 0.0039 | 0.0611 | 0.4248 | 4.7020 | 0.1918 | 1.3606 | 5.5219 | 511.00 | 476.49 | 465.22 | 444.08 |
| Netflix | 0.0023 | 0.0032 | 0.0469 | 0.3363 | 3.7644 | 0.1451 | 1.0387 | 3.5993 | 592.78 | 574.97 | 524.26 | 499.10 |
| Jester | 0.0039 | 0.0053 | 0.1062 | 0.7611 | 8.0272 | 0.3558 | 2.8065 | 14.347 | 587.89 | 577.60 | 536.35 | 529.12 |
| Accidents | 0.0041 | 0.0055 | 0.0408 | 0.2987 | 2.7260 | 0.2533 | 1.6691 | 14.452 | 88.367 | 8.9792 | 2.3818 | 2.4556 |
| Tretail | 0.0011 | 0.0018 | 0.0537 | 0.3901 | 4.2481 | 0.1394 | 1.1912 | 1.5652 | 600.00 | 600.00 | 600.00 | 600.00 |
| Pumsb_star | 0.0026 | 0.0036 | 0.0198 | 0.1537 | 1.5617 | 0.1160 | 0.7219 | 3.6890 | 227.27 | 21.195 | 3.2429 | 2.8835 |
| DNA | 0.0016 | 0.0024 | 0.0356 | 0.2487 | 3.0984 | 0.1037 | 0.7071 | 1.7008 | 600.00 | 571.55 | 528.68 | 520.58 |
| Kosarek | 0.0021 | 0.0031 | 0.0647 | 0.4146 | 4.1906 | 0.1599 | 1.2491 | 3.0625 | 600.00 | 600.00 | 600.00 | 600.00 |
| MSWeb | 0.0015 | 0.0021 | 0.0729 | 0.6354 | 8.2330 | 0.0966 | 0.6793 | 1.4712 | 600.00 | 600.00 | 600.00 | 600.00 |
| Book | 0.0033 | 0.0045 | 0.4813 | 7.6247 | 227.75 | 0.2738 | 2.5722 | 6.6608 | 600.00 | 600.00 | 600.00 | 600.00 |
| EachMovie | 0.0068 | 0.0096 | 0.8479 | 10.820 | 265.50 | 0.6328 | 5.8420 | 32.209 | 600.00 | 600.00 | 600.00 | 600.00 |
| WebKB | 0.0218 | 0.0333 | 3.9349 | 47.608 | 600.00 | 2.4579 | 23.908 | 470.50 | 600.00 | 600.00 | 600.00 | 600.00 |
| Reuters-52 | 0.0049 | 0.0076 | 0.7412 | 9.3151 | 203.78 | 0.4613 | 4.1215 | 19.040 | 600.00 | 600.00 | 600.00 | 600.00 |
| 20 Newsgrp. | 0.0649 | 0.0850 | 4.7551 | 36.124 | 325.72 | 6.9275 | 69.039 | × | 600.00 | 600.00 | 600.00 | 600.00 |
| BBC | 0.0238 | 0.0325 | 0.9577 | 5.8167 | 43.350 | 2.5887 | 25.617 | 434.51 | 600.00 | 600.00 | 600.00 | 600.00 |
| Ad | 0.0030 | 0.0047 | 0.0521 | 0.4167 | 5.2227 | 0.3421 | 3.3667 | 8.2997 | 600.00 | 496.46 | 57.165 | 39.837 |

Figure 4: Average running time (with a 10-minute time limit for each problem). ×: terminated at the time limit with no output.

| Dataset | BT | NG | BS1 | BS10 | BS100 | KBT10 | KBT100 | AMAP | (MC) | (FC) | (FC+O) | (FC+O+S) |
|---|---|---|---|---|---|---|---|---|---|---|---|---|
| NLTCS | 788 | 716 | 912 | 1000 | 1000 | 1000 | 1000 | 1000 | 1000 | 1000 | 1000 | 1000 |
| MSNBC | 822 | 812 | 956 | 1000 | 1000 | 998 | 1000 | 1000 | 1000 | 1000 | 1000 | 1000 |
| KDDCup 2k | 600 | 374 | 549 | 937 | 999 | 879 | 983 | 997 | 1000 | 1000 | 1000 | 1000 |
| Plants | 816 | 605 | 423 | 902 | 999 | 995 | 1000 | 956 | 1000 | 1000 | 1000 | 1000 |
| Audio | 556 | 290 | 437 | 757 | 978 | 737 | 867 | 955 | 521 | 588 | 877 | 913 |
| Netflix | 425 | 256 | 429 | 789 | 980 | 641 | 802 | 906 | 176 | 259 | 912 | 933 |
| Jester | 504 | 174 | 442 | 736 | 941 | 720 | 842 | 888 | 312 | 348 | 793 | 814 |
| Accidents | 942 | 906 | 742 | 982 | 1000 | 996 | 998 | 997 | 1000 | 1000 | 1000 | 1000 |
| Tretail | 751 | 45 | 835 | 946 | 984 | 880 | 947 | 990 | 404 | 550 | 864 | 870 |
| Pumsb_star | 739 | 656 | 681 | 971 | 1000 | 980 | 999 | 985 | 960 | 1000 | 1000 | 1000 |
| DNA | 613 | 108 | 266 | 481 | 712 | 620 | 641 | 957 | 88 | 189 | 517 | 533 |
| Kosarek | 993 | 980 | 620 | 767 | 977 | 995 | 995 | 998 | 0 | 1 | 991 | 994 |
| MSWeb | 935 | 935 | 715 | 775 | 818 | 1000 | 1000 | 1000 | 1 | 1 | 1000 | 1000 |
| Book | 996 | 995 | 652 | 743 | 797 | 998 | 998 | 1000 | 0 | 0 | 995 | 995 |
| EachMovie | 997 | 993 | 823 | 868 | 896 | 998 | 998 | 1000 | 0 | 0 | 993 | 993 |
| WebKB | 1000 | 995 | 847 | 870 | 7 | 1000 | 1000 | 1000 | 0 | 0 | 995 | 995 |
| Reuters-52 | 1000 | 1000 | 962 | 957 | 968 | 1000 | 1000 | 1000 | 0 | 0 | 1000 | 1000 |
| 20 Newsgrp. | 773 | 327 | 383 | 457 | 506 | 918 | 972 | 0 | 1 | 2 | 554 | 556 |
| BBC | 1000 | 994 | 946 | 955 | 983 | 1000 | 1000 | 1000 | 0 | 0 | 1000 | 1000 |
| Ad | 733 | 647 | 279 | 617 | 956 | 964 | 1000 | 1000 | 0 | 410 | 996 | 999 |

Figure 5: Winning counts

## Conclusion

Theoretically, we defined a new inference problem called MAX and presented linear-time reduction from MAP to MAX. This suggests that we can focus on the much simpler MAX problem when studying MAP inference. We also showed that it is almost impossible to find a practical bound for approximate MAP solvers.

Algorithmically, we presented an exact solver based on exhaustive search with pruning, heuristic, and optimization techniques, and an approximate solver based on finding the top $K$ parse trees of the input SPN. Our comprehensive experiments show that the exact solver is reasonably fast and the approximate solver has better overall performance than existing methods.

## Acknowledgments

This work was supported by the National Natural Science Foundation of China (61503248) and Program of Shanghai Subject Chief Scientist (A type) (No.15XD1502900).

## References

Cheng, W. C.; Kok, S.; Pham, H. V.; Chieu, H. L.; and Chai, K. M. A. 2014. Language modeling with sum-product networks. In *INTERSPEECH*.

Choi, A., and Darwiche, A. 2017. On relaxing determinism in arithmetic circuits. In *ICML*.

Conaty, D.; Mauá, D. D.; and de Campos, C. P. 2017. Approximation complexity of maximum a posteriori inference in sum-product networks. In *UAI*.

Crescenzi, P. 1997. A short guide to approximation preserving reductions. In *CCC*.

Darwiche, A., and Marquis, P. 2002. A knowledge compilation map. *JAIR*.

Darwiche, A. 2001. Decomposable negation normal form. *JACM*.

Darwiche, A. 2003. A differential approach to inference in Bayesian networks. *JACM*.

De Campos, C. P. 2011. New complexity results for MAP in Bayesian networks. In *IJCAI*.

Gens, R., and Domingos, P. 2012. Discriminative learning of sum-product networks. In *NIPS*.

Gens, R., and Domingos, P. 2013. Learning the structure of sum-product networks. In *ICML*.

Huang, J.; Chavira, M.; and Darwiche, A. 2006. Solving MAP exactly by searching on compiled arithmetic circuits. In *AAAI*.

Lowd, D., and Domingos, P. 2008. Learning arithmetic circuits. In *UAI*.

Park, J. D. 2002. MAP complexity results and approximation methods. In *UAI*.

Peharz, R.; Kapeller, G.; Mowlaee, P.; and Pernkopf, F. 2014. Modeling speech with sum-product networks: Application to bandwidth extension. In *ICASSP*.

Peharz, R.; Tschiatschek, S.; Pernkopf, F.; Domingos, P. M.; and BioTechMed-Graz, B. 2015. On theoretical properties of sum-product networks. In *AISTATS*.

Peharz, R.; Gens, R.; Pernkopf, F.; and Domingos, P. 2016. On the latent variable interpretation in sum-product networks. *TPAMI*.

Peharz, R.; Gens, R.; and Domingos, P. 2014. Learning selective sum-product networks. In *ICML Workshop on Learning Tractable Probabilistic Models*.

Peharz, D.-I. R. 2015. *Foundations of sum-product networks for probabilistic modeling*. Ph.D. Dissertation, Aalborg University.

Poon, H., and Domingos, P. 2011. Sum-product networks: A new deep architecture. In *UAI*.

Rooshenas, A., and Lowd, D. 2014. Learning sum-product networks with direct and indirect variable interactions. In *ICML*.

Tu, K. 2016. Stochastic And-Or grammars: A unified framework and logic perspective. In *IJCAI*.

Zhao, H.; Melibari, M.; and Poupart, P. 2015. On the relationship between sum-product networks and Bayesian networks. In *ICML*.

Zhao, H.; Poupart, P.; and Gordon, G. 2016. A unified approach for learning the parameters of sum-product networks. In *NIPS*.

## Proof of Lemma 1

**Lemma 1.** *Given a tree-structured BN $\mathcal{B}$, we can construct an SPN $\mathcal{S}$ representing the same distribution with size $|\mathcal{S}| \in \mathcal{O}(|\mathcal{B}|)$ in linear time.*

*Proof.* As mentioned in (Zhao, Melibari, and Poupart 2015), given a BN, there exists an algorithm[1] that builds an SPN representing the same distribution. We adapt their algorithm for tree-structured BNs (Algorithm 4: $BN2SPN$). Given a tree-structured BN $\mathcal{B}$, we would like to show that $|\mathcal{S}| \in \mathcal{O}(|\mathcal{B}|)$ where $\mathcal{S} = BN2SPN(\mathcal{B})$. We denote the numbers of weights, sum nodes, sum arcs, product arcs in $\mathcal{S}$ as $|\boldsymbol{w}|, |V_{\mathsf{S}}|, |A_{\mathsf{S}}|, |A_{\mathsf{P}}|$, and denote the number of parameters and the number of all the values of all the variables in $\mathcal{B}$ as $\#P$, $\#x$. Since we use a map caching result of BUILDSPN$(\mathsf{N}', x')$, we go through Line 7-14 at most $\#x$ times. At Line 3 and 12, whenever we add a sum arc, we use a different parameter in $\mathcal{B}$, thus $|A_{\mathsf{S}}| \in \mathcal{O}(\#P)$ and $|\boldsymbol{w}| \in \mathcal{O}(\#P)$. According to Line 8 and 10, whenever a product arc is added, it links to an indicator or a sum node, thus $|A_{\mathsf{P}}| \in \mathcal{O}(\#x + |V_{\mathsf{S}}|) \subseteq \mathcal{O}(\#x + \#P) = \mathcal{O}(\#P)$. Therefore, $|\mathcal{S}| \in \mathcal{O}(|\boldsymbol{w}| + |A_{\mathsf{S}}| + |A_{\mathsf{P}}|) \subseteq \mathcal{O}(\#P) \subseteq \mathcal{O}(|\mathcal{B}|)$, i.e. $|\mathcal{S}| \in \mathcal{O}(|\mathcal{B}|)$. Furthermore, the algorithm runs in linear time. □

## Experimental results

Table 1 shows the statistics of the SPNs. Figure 7–15 show the results with three additional Q/E/H proportions: 3/7/0, 3/5/2 and 3/1/6.

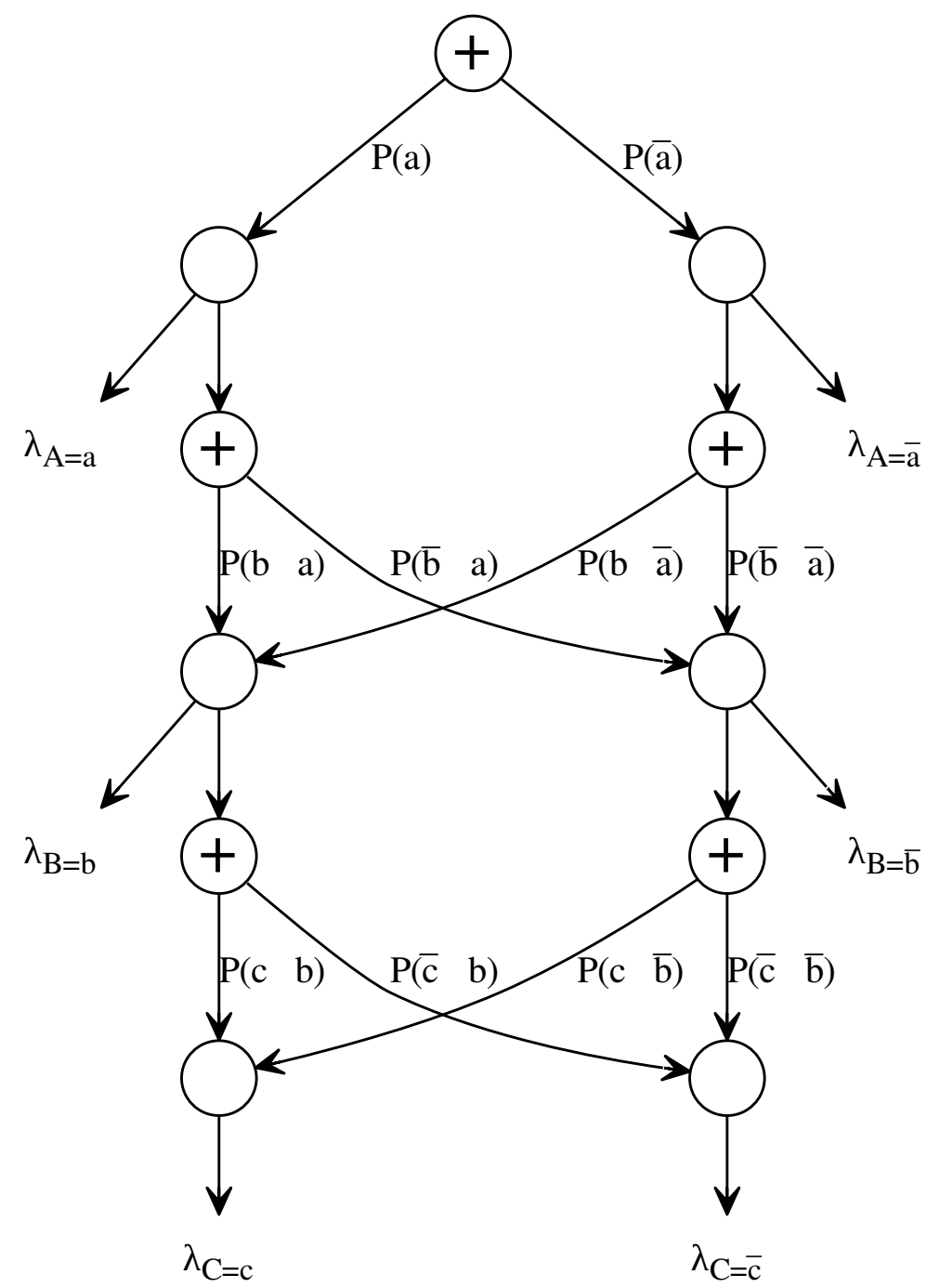

Figure 6: An example of the transferring. The original BN is "$A \to B \to C$" with binary variables.

[1] See also http://spn.cs.washington.edu/faq.shtml

**Algorithm 4** Calculate $\mathcal{S} = BN2SPN(\mathcal{B})$

create sum node $\mathsf{R}$ as the root
**for all** $x \in X_{\mathsf{R}'}$ **do** ▷ $X_{\mathsf{R}'}$ corresponding to the root $\mathsf{R}'$ of $\mathcal{B}$
    add BUILDSPN$(\mathsf{R}', x)$ as child of $\mathsf{R}$ with weight $P(x)$

**function** BUILDSPN$(\mathsf{N}', x')$ ▷ node $\mathsf{N}'$ in $\mathcal{B}$; value $x'$ corresponding to $\mathsf{N}'$
    **if** key $(\mathsf{N}', x')$ exists in $\mathbf{M}$ **then** ▷ $\mathbf{M}$ is a global map caching results
        **return** $\mathbf{M}(\mathsf{N}', x')$
    create a product node $\mathsf{P}$
    add $\lambda_{X_{\mathsf{N}'}=x'}$ as child of $\mathsf{P}$ ▷ $X_{\mathsf{N}'}$ corresponding to $\mathsf{N}'$
    **for all** $\mathsf{C}' \in \mathbf{ch}(\mathsf{N}')$ **do**
        create a sum node $\mathsf{S}$ as child of $\mathsf{P}$
        **for all** $x \in \mathbf{val}(X_{\mathsf{C}'})$ **do** ▷ $X_{\mathsf{C}'}$ corresponding to $\mathsf{C}'$
            add BUILDSPN$(\mathsf{C}', x)$ as child of $\mathsf{S}$ with weight $P(x \mid x')$
    store $(\mathsf{N}', e) \rightarrow \mathsf{P}$ in map $\mathbf{M}$
    **return** $\mathsf{P}$

| **Dataset** | **#Vars** | **#Arcs** | **Dataset** | **#Vars** | **#Arcs** |
|---|---|---|---|---|---|
| NLTCS | 16 | 6471 | DNA | 180 | 51843 |
| MSNBC | 17 | 59835 | Kosarek | 190 | 73816 |
| KDDCup 2k | 65 | 132992 | MSWeb | 294 | 47630 |
| Plants | 69 | 115384 | Book | 500 | 115309 |
| Audio | 100 | 95587 | EachMovie | 500 | 254922 |
| Netflix | 100 | 74510 | WebKB | 839 | 959922 |
| Jester | 100 | 160868 | Reuters-52 | 889 | 194574 |
| Accidents | 111 | 147599 | 20 Newsgrp. | 910 | 2598116 |
| Tretail | 135 | 54469 | BBC | 1058 | 937697 |
| Pumsb_star | 163 | 70960 | Ad | 1556 | 126831 |

Table 1: The statistics of the SPNs learned from the 20 datasets. #Vars denotes the number of variables and #Arcs denotes the number of arcs in the learned SPN.

| Dataset | MC | FC | FC+O | FC+O+S |
|---|---|---|---|---|
| NLTCS | 1000 | 1000 | 1000 | 1000 |
| MSNBC | 1000 | 1000 | 1000 | 1000 |
| KDDCup 2k | 1000 | 1000 | 1000 | 1000 |
| Plants | 1000 | 1000 | 1000 | 1000 |
| Audio | 27 | 40 | 58 | 68 |
| Netflix | 12 | 37 | 74 | 81 |
| Jester | 5 | 10 | 21 | 21 |
| Accidents | 1000 | 1000 | 1000 | 1000 |
| Pumsb_star | 930 | 1000 | 1000 | 1000 |
| DNA | 163 | 469 | 528 | 646 |
| Ad | 0 | 253 | 926 | 984 |

Figure 7: Finishing counts of exact solvers where the Q/E/H proportion is 3/7/0. We skip the rows of all zeros.

| Dataset | BT | NG | BS1 | BS10 | BS100 | KBT10 | KBT100 | AMAP | (MC) | (FC) | (FC+O) | (FC+O+S) |
|---|---|---|---|---|---|---|---|---|---|---|---|---|
| NLTCS | 0.0003 | 0.0003 | 0.0013 | 0.0088 | 0.0315 | 0.0146 | 0.0401 | 0.0578 | 0.0059 | 0.0103 | 0.0085 | 0.0068 |
| MSNBC | 0.0019 | 0.0023 | 0.0099 | 0.0933 | 0.3958 | 0.0938 | 0.3008 | 3.8286 | 0.0229 | 0.0466 | 0.0404 | 0.0336 |
| KDDCup 2k | 0.0030 | 0.0043 | 0.0468 | 0.3635 | 3.4883 | 0.3156 | 2.3678 | 10.344 | 9.1801 | 10.4022 | 8.9809 | 6.5303 |
| Plants | 0.0025 | 0.0039 | 0.0294 | 0.2235 | 2.1137 | 0.2298 | 1.6536 | 7.9471 | 2.1953 | 1.7454 | 0.6624 | 0.5646 |
| Audio | 0.0027 | 0.0035 | 0.0660 | 0.4752 | 5.2951 | 0.1964 | 1.3801 | 5.5403 | 592.68 | 587.59 | 583.86 | 582.09 |
| Netflix | 0.0026 | 0.0032 | 0.0522 | 0.3679 | 4.2920 | 0.1468 | 1.0520 | 3.6339 | 597.71 | 591.62 | 580.93 | 579.27 |
| Jester | 0.0038 | 0.0054 | 0.1106 | 0.7993 | 8.6237 | 0.3664 | 2.8575 | 14.388 | 599.06 | 597.31 | 593.90 | 593.95 |
| Accidents | 0.0038 | 0.0054 | 0.0401 | 0.2786 | 2.4866 | 0.2601 | 1.6897 | 14.516 | 62.074 | 7.4015 | 2.3775 | 2.9021 |
| Tretail | 0.0012 | 0.0017 | 0.0536 | 0.3913 | 4.1879 | 0.1432 | 1.2005 | 1.5676 | 600.00 | 600.00 | 600.00 | 600.00 |
| Pumsb_star | 0.0028 | 0.0035 | 0.0212 | 0.1556 | 1.5671 | 0.1192 | 0.7217 | 3.6697 | 235.81 | 23.758 | 4.5351 | 4.1720 |
| DNA | 0.0016 | 0.0025 | 0.0261 | 0.1851 | 2.1882 | 0.1068 | 0.7150 | 1.6997 | 555.31 | 420.43 | 356.78 | 306.56 |
| Kosarek | 0.0023 | 0.0032 | 0.0648 | 0.4094 | 4.1089 | 0.1633 | 1.2668 | 3.0685 | 600.00 | 600.00 | 600.00 | 600.00 |
| MSWeb | 0.0015 | 0.0022 | 0.0720 | 0.6217 | 8.1955 | 0.0977 | 0.6750 | 1.4410 | 600.00 | 600.00 | 600.00 | 600.00 |
| Book | 0.0033 | 0.0050 | 0.4802 | 7.6135 | 233.34 | 0.2785 | 2.5776 | 6.6537 | 600.00 | 600.00 | 600.00 | 600.00 |
| EachMovie | 0.0081 | 0.0102 | 0.8590 | 11.271 | 275.82 | 0.6474 | 5.8719 | 32.295 | 600.00 | 600.00 | 600.00 | 600.00 |
| WebKB | 0.0224 | 0.0322 | 3.9633 | 47.378 | 600.00 | 2.4882 | 24.170 | 470.08 | 600.00 | 600.00 | 600.00 | 600.00 |
| Reuters-52 | 0.0050 | 0.0074 | 0.7438 | 9.8599 | 207.21 | 0.4679 | 4.1219 | 19.026 | 600.00 | 600.00 | 600.00 | 600.00 |
| 20 Newsgrp. | 0.0615 | 0.0898 | 5.0757 | 40.059 | 377.18 | 6.9467 | 69.271 | × | 600.00 | 600.00 | 600.00 | 600.00 |
| BBC | 0.0232 | 0.0320 | 0.9524 | 5.8811 | 43.137 | 2.5735 | 25.816 | 432.35 | 600.00 | 600.00 | 600.00 | 600.00 |
| Ad | 0.0033 | 0.0050 | 0.0564 | 0.4393 | 5.6647 | 0.3452 | 3.4205 | 8.2793 | 600.00 | 538.58 | 118.45 | 41.213 |

Figure 8: Average running time (with a 10-minute time limit for each problem) where the Q/E/H proportion is 3/7/0. $\times$: terminated at the time limit with no output.

| Dataset | BT | NG | BS1 | BS10 | BS100 | KBT10 | KBT100 | AMAP | (MC) | (FC) | (FC+O) | (FC+O+S) |
|---|---|---|---|---|---|---|---|---|---|---|---|---|
| NLTCS | 818 | 737 | 930 | 1000 | 1000 | 1000 | 1000 | 1000 | 1000 | 1000 | 1000 | 1000 |
| MSNBC | 787 | 714 | 919 | 1000 | 1000 | 998 | 1000 | 1000 | 1000 | 1000 | 1000 | 1000 |
| KDDCup 2k | 738 | 726 | 559 | 951 | 1000 | 971 | 996 | 1000 | 1000 | 1000 | 1000 | 1000 |
| Plants | 829 | 799 | 645 | 961 | 1000 | 997 | 1000 | 984 | 1000 | 1000 | 1000 | 1000 |
| Audio | 756 | 368 | 393 | 595 | 918 | 883 | 968 | 914 | 124 | 151 | 738 | 744 |
| Netflix | 814 | 401 | 423 | 656 | 935 | 929 | 975 | 950 | 79 | 121 | 790 | 800 |
| Jester | 688 | 306 | 384 | 638 | 927 | 849 | 949 | 876 | 122 | 132 | 717 | 724 |
| Accidents | 998 | 996 | 889 | 999 | 1000 | 1000 | 1000 | 1000 | 1000 | 1000 | 1000 | 1000 |
| Tretail | 373 | 65 | 924 | 977 | 989 | 510 | 694 | 980 | 338 | 604 | 986 | 986 |
| Pumsb_star | 910 | 824 | 795 | 980 | 1000 | 999 | 1000 | 997 | 937 | 1000 | 1000 | 1000 |
| DNA | 833 | 539 | 544 | 766 | 918 | 889 | 942 | 985 | 286 | 617 | 874 | 919 |
| Kosarek | 998 | 1000 | 779 | 871 | 995 | 1000 | 1000 | 1000 | 0 | 2 | 1000 | 1000 |
| MSWeb | 979 | 979 | 818 | 873 | 917 | 1000 | 1000 | 1000 | 0 | 0 | 1000 | 1000 |
| Book | 1000 | 1000 | 761 | 857 | 885 | 1000 | 1000 | 1000 | 0 | 0 | 1000 | 1000 |
| EachMovie | 1000 | 1000 | 886 | 920 | 954 | 1000 | 1000 | 1000 | 0 | 0 | 1000 | 1000 |
| WebKB | 1000 | 1000 | 911 | 931 | 9 | 1000 | 1000 | 1000 | 0 | 0 | 1000 | 1000 |
| Reuters-52 | 1000 | 1000 | 964 | 965 | 979 | 1000 | 1000 | 1000 | 0 | 0 | 1000 | 1000 |
| 20 Newsgrp. | 899 | 674 | 653 | 686 | 735 | 959 | 986 | 0 | 0 | 0 | 845 | 851 |
| BBC | 1000 | 998 | 966 | 978 | 991 | 1000 | 1000 | 1000 | 0 | 0 | 1000 | 1000 |
| Ad | 899 | 856 | 449 | 799 | 994 | 991 | 1000 | 1000 | 0 | 277 | 1000 | 1000 |

Figure 9: Winning counts where the Q/E/H proportion is 3/7/0.

| Dataset | MC | FC | FC+O | FC+O+S |
|---|---|---|---|---|
| NLTCS | 1000 | 1000 | 1000 | 1000 |
| MSNBC | 1000 | 1000 | 1000 | 1000 |
| KDDCup 2k | 1000 | 1000 | 1000 | 1000 |
| Plants | 1000 | 1000 | 1000 | 1000 |
| Audio | 56 | 92 | 112 | 126 |
| Netflix | 17 | 53 | 139 | 171 |
| Jester | 10 | 19 | 50 | 53 |
| Accidents | 1000 | 1000 | 1000 | 1000 |
| Pumsb_star | 943 | 1000 | 1000 | 1000 |
| DNA | 78 | 245 | 287 | 381 |
| Ad | 0 | 297 | 929 | 981 |

Figure 10: Finishing counts of exact solvers where the Q/E/H proportion is 3/5/2. We skip the rows of all zeros.

| Dataset | BT | NG | BS1 | BS10 | BS100 | KBT10 | KBT100 | AMAP | (MC) | (FC) | (FC+O) | (FC+O+S) |
|---|---|---|---|---|---|---|---|---|---|---|---|---|
| NLTCS | 0.0003 | 0.0004 | 0.0013 | 0.0086 | 0.0309 | 0.0149 | 0.0398 | 0.0583 | 0.0057 | 0.0100 | 0.0084 | 0.0068 |
| MSNBC | 0.0017 | 0.0024 | 0.0097 | 0.0879 | 0.3681 | 0.0917 | 0.3004 | 3.8341 | 0.0210 | 0.0452 | 0.0397 | 0.0328 |
| KDDCup 2k | 0.0030 | 0.0043 | 0.0474 | 0.3791 | 3.6987 | 0.3101 | 2.3645 | 10.352 | 9.9253 | 11.606 | 10.580 | 7.5813 |
| Plants | 0.0027 | 0.0039 | 0.0294 | 0.2309 | 2.2426 | 0.2253 | 1.6677 | 7.9445 | 2.3315 | 1.8092 | 0.7212 | 0.6128 |
| Audio | 0.0024 | 0.0037 | 0.0655 | 0.4656 | 5.2685 | 0.1914 | 1.3786 | 5.5393 | 583.11 | 571.47 | 565.40 | 560.80 |
| Netflix | 0.0026 | 0.0032 | 0.0502 | 0.3500 | 4.0907 | 0.1444 | 1.0511 | 3.6458 | 600.00 | 587.04 | 566.21 | 558.30 |
| Jester | 0.0036 | 0.0052 | 0.1095 | 0.7954 | 8.6719 | 0.3577 | 2.8446 | 14.321 | 600.00 | 594.47 | 587.02 | 585.81 |
| Accidents | 0.0037 | 0.0055 | 0.0412 | 0.2843 | 2.5304 | 0.2538 | 1.6956 | 14.588 | 73.553 | 7.8473 | 2.4200 | 2.8285 |
| Tretail | 0.0011 | 0.0018 | 0.0538 | 0.3965 | 4.1831 | 0.1396 | 1.2108 | 1.5648 | 600.00 | 600.00 | 600.00 | 600.00 |
| Pumsb_star | 0.0028 | 0.0033 | 0.0203 | 0.1569 | 1.5607 | 0.1167 | 0.7407 | 3.6831 | 231.90 | 22.910 | 4.1228 | 3.7466 |
| DNA | 0.0016 | 0.0025 | 0.0300 | 0.2152 | 2.6148 | 0.1035 | 0.7203 | 1.6965 | 580.79 | 511.21 | 466.81 | 433.07 |
| Kosarek | 0.0022 | 0.0032 | 0.0648 | 0.4088 | 4.1708 | 0.1595 | 1.2727 | 3.0606 | 600.00 | 600.00 | 600.00 | 600.00 |
| MSWeb | 0.0014 | 0.0023 | 0.0718 | 0.6321 | 8.1353 | 0.0950 | 0.6818 | 1.4327 | 600.00 | 600.00 | 600.00 | 600.00 |
| Book | 0.0037 | 0.0048 | 0.4816 | 7.5783 | 231.10 | 0.2719 | 2.6041 | 6.6537 | 600.00 | 600.00 | 600.00 | 600.00 |
| EachMovie | 0.0069 | 0.0102 | 0.8506 | 11.150 | 275.42 | 0.6272 | 5.8483 | 32.032 | 600.00 | 600.00 | 600.00 | 600.00 |
| WebKB | 0.0230 | 0.0333 | 3.9728 | 47.812 | 600.00 | 2.4572 | 24.208 | 472.37 | 600.00 | 600.00 | 600.00 | 600.00 |
| Reuters-52 | 0.0048 | 0.0074 | 0.7419 | 9.6247 | 204.45 | 0.4699 | 4.1057 | 19.024 | 600.00 | 600.00 | 600.00 | 600.00 |
| 20 Newsgrp. | 0.0617 | 0.0867 | 5.0289 | 39.051 | 359.49 | 6.9670 | 69.570 | × | 600.00 | 600.00 | 600.00 | 600.00 |
| BBC | 0.0222 | 0.0330 | 0.9442 | 5.8011 | 43.348 | 2.5595 | 25.868 | 436.20 | 600.00 | 600.00 | 600.00 | 600.00 |
| Ad | 0.0031 | 0.0048 | 0.0553 | 0.4253 | 5.4803 | 0.3403 | 3.3809 | 8.2532 | 600.00 | 522.30 | 122.62 | 42.105 |

Figure 11: Average running time (with a 10-minute time limit for each problem) where the Q/E/H proportion is 3/5/2. $\times$: terminated at the time limit with no output.

| Dataset | BT | NG | BS1 | BS10 | BS100 | KBT10 | KBT100 | AMAP | (MC) | (FC) | (FC+O) | (FC+O+S) |
|---|---|---|---|---|---|---|---|---|---|---|---|---|
| NLTCS | 802 | 721 | 931 | 1000 | 1000 | 1000 | 1000 | 1000 | 1000 | 1000 | 1000 | 1000 |
| MSNBC | 783 | 743 | 939 | 1000 | 1000 | 999 | 1000 | 1000 | 1000 | 1000 | 1000 | 1000 |
| KDDCup 2k | 656 | 590 | 544 | 957 | 1000 | 928 | 988 | 998 | 1000 | 1000 | 1000 | 1000 |
| Plants | 809 | 734 | 536 | 945 | 1000 | 997 | 1000 | 976 | 1000 | 1000 | 1000 | 1000 |
| Audio | 630 | 304 | 403 | 673 | 948 | 791 | 903 | 957 | 242 | 282 | 786 | 809 |
| Netflix | 615 | 330 | 451 | 751 | 971 | 799 | 915 | 946 | 97 | 174 | 863 | 886 |
| Jester | 589 | 239 | 409 | 689 | 935 | 813 | 914 | 877 | 204 | 239 | 759 | 776 |
| Accidents | 987 | 981 | 876 | 993 | 1000 | 1000 | 1000 | 998 | 1000 | 1000 | 1000 | 1000 |
| Tretail | 543 | 40 | 874 | 958 | 982 | 685 | 832 | 984 | 356 | 561 | 949 | 950 |
| Pumsb_star | 810 | 725 | 730 | 971 | 1000 | 985 | 1000 | 988 | 947 | 1000 | 1000 | 1000 |
| DNA | 555 | 277 | 425 | 647 | 878 | 601 | 697 | 970 | 179 | 388 | 760 | 797 |
| Kosarek | 998 | 994 | 702 | 841 | 992 | 999 | 1000 | 999 | 0 | 0 | 1000 | 1000 |
| MSWeb | 973 | 972 | 779 | 870 | 891 | 1000 | 1000 | 1000 | 0 | 1 | 1000 | 1000 |
| Book | 1000 | 1000 | 725 | 820 | 864 | 1000 | 1000 | 1000 | 0 | 0 | 1000 | 1000 |
| EachMovie | 1000 | 1000 | 871 | 911 | 939 | 1000 | 1000 | 1000 | 0 | 0 | 1000 | 1000 |
| WebKB | 1000 | 1000 | 875 | 914 | 5 | 1000 | 1000 | 1000 | 0 | 0 | 1000 | 1000 |
| Reuters-52 | 1000 | 1000 | 948 | 964 | 975 | 1000 | 1000 | 1000 | 0 | 0 | 1000 | 1000 |
| 20 Newsgrp. | 810 | 513 | 546 | 576 | 663 | 929 | 976 | 0 | 0 | 0 | 738 | 740 |
| BBC | 999 | 992 | 955 | 965 | 988 | 1000 | 1000 | 1000 | 0 | 0 | 1000 | 1000 |
| Ad | 851 | 784 | 345 | 700 | 983 | 983 | 999 | 1000 | 0 | 321 | 993 | 1000 |

Figure 12: Winning counts where the Q/E/H proportion is 3/5/2.

| Dataset | MC | FC | FC+O | FC+O+S |
|---|---|---|---|---|
| NLTCS | 1000 | 1000 | 1000 | 1000 |
| MSNBC | 1000 | 1000 | 1000 | 1000 |
| KDDCup 2k | 1000 | 1000 | 1000 | 1000 |
| Plants | 1000 | 1000 | 1000 | 1000 |
| Audio | 838 | 880 | 850 | 909 |
| Netflix | 30 | 132 | 468 | 628 |
| Jester | 250 | 399 | 909 | 876 |
| Accidents | 1000 | 1000 | 1000 | 1000 |
| Tretail | 383 | 421 | 514 | 486 |
| Pumsb_star | 963 | 1000 | 1000 | 1000 |
| DNA | 13 | 53 | 291 | 277 |
| Kosarak | 67 | 118 | 189 | 161 |
| Ad | 2 | 770 | 997 | 1000 |

Figure 13: Finishing counts of exact solvers where the Q/E/H proportion is 3/1/6. We skip the rows of all zeros.

| Dataset | BT | NG | BS1 | BS10 | BS100 | KBT10 | KBT100 | AMAP | (MC) | (FC) | (FC+O) | (FC+O+S) |
|---|---|---|---|---|---|---|---|---|---|---|---|---|
| NLTCS | 0.0003 | 0.0004 | 0.0012 | 0.0079 | 0.0301 | 0.0151 | 0.0409 | 0.0587 | 0.0046 | 0.0087 | 0.0085 | 0.0064 |
| MSNBC | 0.0018 | 0.0024 | 0.0084 | 0.0540 | 0.2340 | 0.0959 | 0.3030 | 3.8305 | 0.0147 | 0.0387 | 0.0384 | 0.0315 |
| KDDCup 2k | 0.0029 | 0.0042 | 0.0400 | 0.3601 | 4.0158 | 0.3136 | 2.3791 | 10.351 | 1.3798 | 1.7602 | 1.5217 | 1.1812 |
| Plants | 0.0027 | 0.0038 | 0.0251 | 0.3115 | 4.0004 | 0.2296 | 1.6599 | 7.9683 | 1.0144 | 0.9146 | 0.4174 | 0.3264 |
| Audio | 0.0027 | 0.0039 | 0.0553 | 0.3791 | 4.1864 | 0.1955 | 1.4052 | 5.5822 | 157.62 | 128.37 | 148.10 | 127.98 |
| Netflix | 0.0021 | 0.0031 | 0.0450 | 0.3369 | 3.7665 | 0.1439 | 1.0626 | 3.6299 | 593.58 | 571.87 | 445.91 | 392.27 |
| Jester | 0.0037 | 0.0054 | 0.0983 | 0.6742 | 7.0332 | 0.3553 | 2.8626 | 14.335 | 509.29 | 461.44 | 173.86 | 215.89 |
| Accidents | 0.0039 | 0.0056 | 0.0430 | 0.3856 | 3.8762 | 0.2541 | 1.6802 | 14.558 | 139.09 | 13.458 | 2.1965 | 2.9141 |
| Tretail | 0.0012 | 0.0019 | 0.0493 | 0.3496 | 3.8417 | 0.1394 | 1.2117 | 1.5719 | 405.92 | 385.78 | 334.24 | 354.89 |
| Pumsb_star | 0.0027 | 0.0036 | 0.0194 | 0.1585 | 1.6159 | 0.1169 | 0.7421 | 3.6900 | 232.07 | 20.570 | 1.7620 | 2.0562 |
| DNA | 0.0016 | 0.0024 | 0.0413 | 0.2759 | 3.4332 | 0.1031 | 0.7223 | 1.6992 | 600.00 | 583.84 | 446.43 | 463.16 |
| Kosarek | 0.0021 | 0.0032 | 0.0635 | 0.4054 | 4.3498 | 0.1590 | 1.2499 | 3.0662 | 574.16 | 547.07 | 514.14 | 527.67 |
| MSWeb | 0.0016 | 0.0022 | 0.0731 | 0.6416 | 8.3179 | 0.0960 | 0.6958 | 1.4577 | 600.00 | 600.00 | 600.00 | 600.00 |
| Book | 0.0033 | 0.0046 | 0.4867 | 7.5690 | 234.99 | 0.2697 | 2.6208 | 6.6607 | 600.00 | 600.00 | 600.00 | 600.00 |
| EachMovie | 0.0069 | 0.0099 | 0.8444 | 10.581 | 247.61 | 0.6290 | 5.9196 | 32.054 | 600.00 | 600.00 | 600.00 | 600.00 |
| WebKB | 0.0213 | 0.0318 | 3.8961 | 47.464 | 600.00 | 2.4693 | 24.285 | 472.98 | 600.00 | 600.00 | 600.00 | 600.00 |
| Reuters-52 | 0.0050 | 0.0071 | 0.7505 | 8.9079 | 200.30 | 0.4675 | 4.1760 | 18.982 | 600.00 | 600.00 | 600.00 | 600.00 |
| 20 Newsgrp. | 0.0666 | 0.0839 | 4.0068 | 27.426 | 224.48 | 6.9801 | 70.094 | × | 600.00 | 600.00 | 600.00 | 600.00 |
| BBC | 0.0228 | 0.0328 | 0.9475 | 5.8023 | 42.986 | 2.5851 | 26.122 | 436.97 | 600.00 | 600.00 | 600.00 | 600.00 |
| Ad | 0.0032 | 0.0048 | 0.0452 | 0.3633 | 4.4554 | 0.3444 | 3.3776 | 8.2731 | 600.00 | 317.66 | 15.699 | 20.320 |

Figure 14: Average running time (with a 10-minute time limit for each problem) where the Q/E/H proportion is 3/1/6. ×: terminated at the time limit with no output.

| Dataset | BT | NG | BS1 | BS10 | BS100 | KBT10 | KBT100 | AMAP | (MC) | (FC) | (FC+O) | (FC+O+S) |
|---|---|---|---|---|---|---|---|---|---|---|---|---|
| NLTCS | 855 | 859 | 846 | 1000 | 1000 | 1000 | 1000 | 1000 | 1000 | 1000 | 1000 | 1000 |
| MSNBC | 821 | 763 | 995 | 1000 | 1000 | 1000 | 1000 | 1000 | 1000 | 1000 | 1000 | 1000 |
| KDDCup 2k | 800 | 540 | 762 | 965 | 999 | 987 | 1000 | 1000 | 1000 | 1000 | 1000 | 1000 |
| Plants | 699 | 445 | 427 | 935 | 999 | 977 | 1000 | 970 | 1000 | 1000 | 1000 | 1000 |
| Audio | 699 | 105 | 558 | 871 | 988 | 878 | 943 | 983 | 904 | 927 | 932 | 970 |
| Netflix | 372 | 124 | 335 | 729 | 971 | 598 | 753 | 886 | 207 | 338 | 892 | 934 |
| Jester | 650 | 84 | 521 | 851 | 994 | 857 | 939 | 958 | 397 | 514 | 960 | 946 |
| Accidents | 551 | 475 | 417 | 893 | 996 | 848 | 974 | 966 | 1000 | 1000 | 1000 | 1000 |
| Tretail | 849 | 149 | 915 | 973 | 996 | 994 | 1000 | 1000 | 723 | 877 | 981 | 981 |
| Pumsb_star | 581 | 448 | 496 | 931 | 1000 | 893 | 992 | 976 | 967 | 1000 | 1000 | 1000 |
| DNA | 990 | 54 | 79 | 239 | 537 | 992 | 993 | 999 | 22 | 73 | 305 | 293 |
| Kosarek | 654 | 303 | 216 | 329 | 558 | 806 | 871 | 985 | 152 | 296 | 868 | 874 |
| MSWeb | 812 | 791 | 487 | 615 | 636 | 999 | 1000 | 997 | 10 | 10 | 1000 | 1000 |
| Book | 644 | 3 | 36 | 31 | 28 | 935 | 988 | 1000 | 592 | 666 | 208 | 208 |
| EachMovie | 540 | 138 | 103 | 125 | 134 | 659 | 759 | 914 | 1 | 7 | 244 | 244 |
| WebKB | 916 | 65 | 59 | 62 | 6 | 944 | 957 | 997 | 1 | 1 | 128 | 128 |
| Reuters-52 | 982 | 826 | 745 | 760 | 781 | 993 | 999 | 1000 | 0 | 0 | 896 | 900 |
| 20 Newsgrp. | 975 | 332 | 341 | 446 | 536 | 991 | 1000 | 0 | 0 | 0 | 400 | 400 |
| BBC | 998 | 986 | 894 | 915 | 980 | 1000 | 1000 | 1000 | 0 | 0 | 999 | 999 |
| Ad | 782 | 643 | 327 | 740 | 972 | 986 | 999 | 1000 | 3 | 789 | 1000 | 1000 |

Figure 15: Winning counts where the Q/E/H proportion is 3/1/6.